\documentclass[10pt,letterpaper]{article}

\usepackage{cogsci}

\cogscifinalcopy 

\usepackage{pslatex}
\usepackage{apacite}
\usepackage{float} 
\usepackage{graphicx}
\usepackage{caption}
\usepackage{booktabs}
\usepackage{comment}
\usepackage{amsmath} 
\usepackage[normalem]{ulem}
\usepackage{ulem}
\usepackage{wrapfig}

\usepackage{hyperref}
\usepackage{url}

\usepackage{natbib}
\title{Visual communication of object concepts at different levels of abstraction}
 
\author{
\textbf{Justin Yang} \\ Dept. of Cognitive Science \\  UC San Diego \\ \texttt{justin-yang@ucsd.edu}
 \And
\textbf{Judith E. Fan} \\ Dept. of Psychology \\ UC San Diego \\ \texttt{jefan@ucsd.edu}
}

\begin{document}

\maketitle 

\begin{abstract}
People can produce drawings of specific entities (e.g., Garfield), as well as general categories (e.g., ``cat'').
What explains this ability to produce such varied drawings of even highly familiar object concepts?
We hypothesized that drawing objects at different levels of abstraction depends on both sensory information and representational goals, such that drawings intended to portray a recently seen object preserve more detail than those intended to represent a category. 
Participants drew objects cued either with a photo or a category label. 
For each cue type, half the participants aimed to draw a specific exemplar; the other half aimed to draw the category.
We found that label-cued category drawings were the most recognizable at the basic level, whereas photo-cued exemplar drawings were the least recognizable.
Together, these findings highlight the importance of task context for explaining how people use drawings to communicate visual concepts in different ways.

\end{abstract}

\section{Introduction}

One of the most distinctive aspects of human communication is that it goes beyond vocal production --- humans have devised many ways to make their ideas both visible and durable. 
From etchings on cave walls to modern digital displays, some of the most significant inventions in human history include technologies that externalize our thoughts in visual form.
Despite the importance of such technologies, little is known about how the human mind is capable of using them in such varied ways. Perhaps the most basic and versatile of these technologies is drawing.

\begin{figure}[t]
\centering
\includegraphics[width=.99\linewidth]{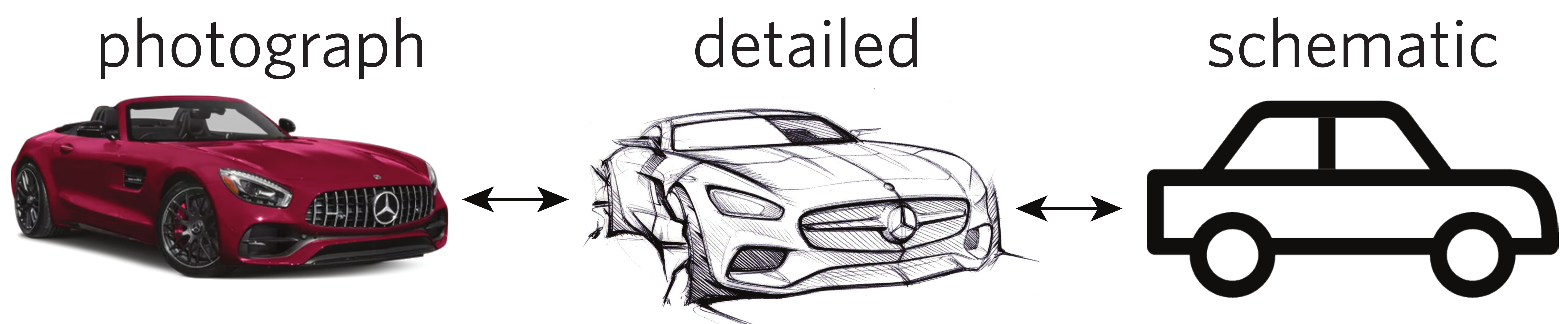}
\caption{Humans use drawings to communicate meanings spanning many levels of abstraction.}
\label{visual_abstraction}
\end{figure}

Drawing predates the invention of writing \citep{clottes2008cave} and is pervasive across many cultures \citep{gombrich1989story}.
It has long inspired scientists to investigate the mental representation of concepts in children \citep{minsky1972artificial,KarmiloffSmith:1990ty} and clinical populations \citep{Bozeat:2003hk,chen2012clock}. 
Despite drawing's importance as a technology for expressing human knowledge, the underlying cognitive mechanisms underpinning our ability to produce such varied drawings are relatively unknown.
In particular, prior work has seldom addressed the question of how drawing enables the flexible expression of meanings across different levels of visual abstraction, ranging from detailed drawings of specific objects to sparse drawings that communicate information about basic-level categories (Fig.~\ref{visual_abstraction}). 
As a consequence, theories of how visual images convey information at different levels of abstraction are comparatively impoverished, by contrast with theories of how such semantic hierarchies are encoded in natural language \citep{miller1995wordnet,xu2007word,rosch1976basic}.


Here we investigate the cognitive and task constraints that enable such flexible expression of visual knowledge. 
Specifically, we explore the hypothesis that the ability to draw objects at different levels of abstraction is jointly dependent on sensory information and representational goals (i.e., the subject of the drawing), such that drawings intended to portray a specific exemplar contain different semantic information than drawings intended to represent a category. 
To test this hypothesis, we conducted a systematic investigation of the semantic information contained in drawings of a wide variety of visual objects using a combination of crowdsourcing and model-based analyses. 

Our approach builds on a growing body of literature using drawing paradigms to investigate various aspects of cognition, including learning \citep{FanYaminsTurkBrowne2018,fiorella2018drawing}, communication \cite{hawkins2019disentangling,fan2020pragmatic}, memory \citep{bainbridge2019drawings,roberts2020drawing}, and development \citep{dillon2020rooms,Long2019}. 
A key limitation of this prior work is that it has generally restricted their focus to drawings produced at a specific level of abstraction by using either category labels \textit{or} natural images as cues, potentially restricting the dynamic range over which drawings can vary. 
To address this limitation, here we directly manipulate sensory information and representational goals within the same paradigm, allowing us to disentangle their contributions to the semantic content of the resulting drawing. 

\begin{figure}[hbtp]
\centering
\includegraphics[width=.8\linewidth]{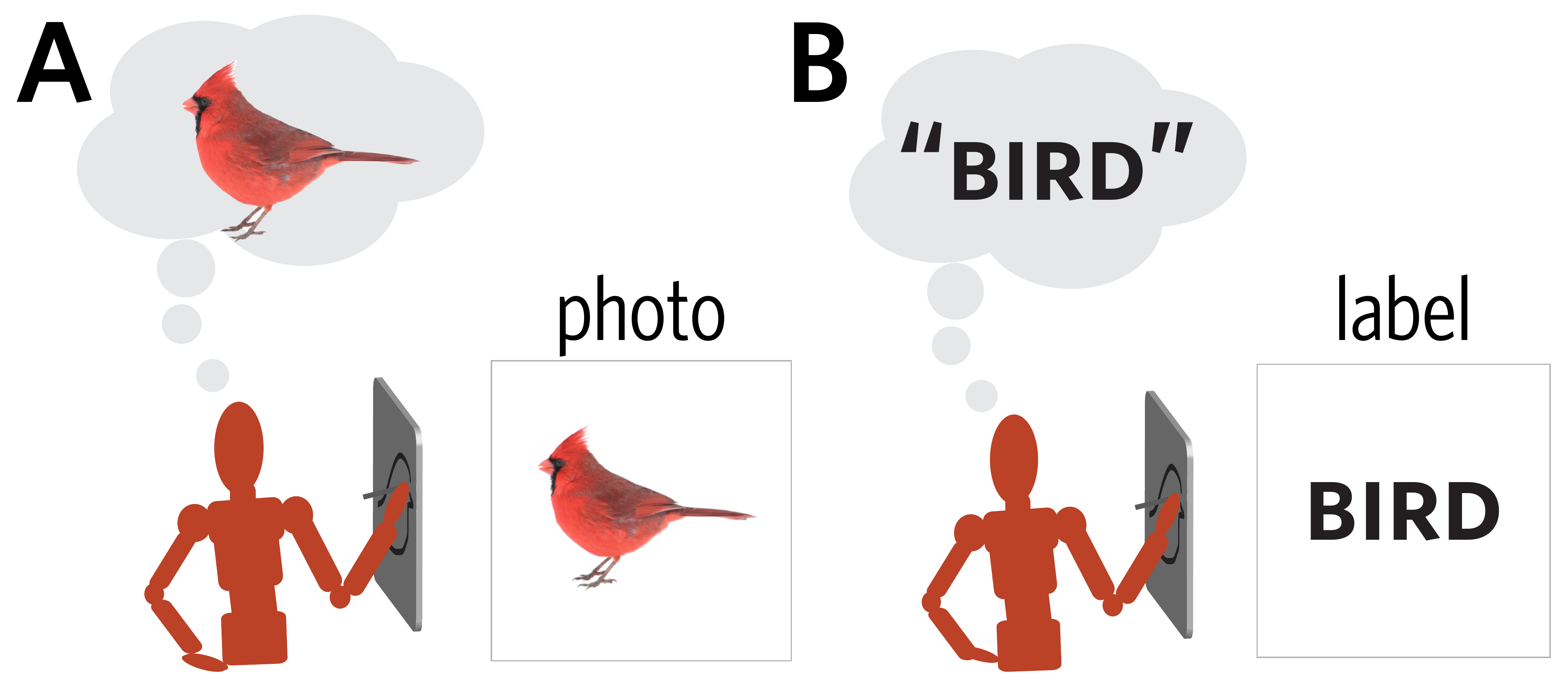}
\caption{Study 1 Task Procedure. A: On photo-cue trials, participants aimed to produce a drawing of the photographed exemplar. B: On label-cue trials, participants aimed to produce a drawing of the labeled category.}
\label{photodraw_task}
\end{figure}

\section{Study 1: How do drawings cued by prototypical exemplars differ from drawings cued by category labels?}

The goal of our first study was to explore the extent to which drawings of objects produced in the presence of a typical exemplar differed from drawings based solely on prior semantic knowledge of that category. 
Towards this end, we manipulated whether participants were cued with a photo of a highly prototypical exemplar or with a category label before producing their drawing (Fig.~\ref{photodraw_task}).
Insofar as the photo provided a visual reminder to participants of the diagnostic properties of each object, we predicted that photo-cued drawings would be easier to recognize at the category level than label-cued ones. 
Alternatively, insofar as category labels more strongly activate information that is diagnostic of basic-level category membership than even photos of typical exemplars, we predicted that \textit{label}-cued drawings would be more recognizable at the category level \citep{lupyan2012evocative}.

\subsection{Methods} 
\subsubsection{Participants} 
57 English-speaking adults recruited via Amazon Mechanical Turk (AMT) completed the study (29 male, 36.8 years). 
Each participant received \$2.00 for their participation (approx. \$12/hr) and provided informed consent as per our institution's IRB.
Data from 4 participants who met our pre-registered criteria were excluded from further analyses.\footnote{Data from an entire session were excluded if it contained at least three blank drawings, at least three `incomplete' drawings consisting of a single stroke, or at least three invalid drawings (containing text, surrounding context, or other inappropriate content).} 

\subsubsection{Stimuli} We obtained 3 color photographs of prototypical exemplars from each of 12 familiar object categories: \textit{airplane, bike, bird, car, cat, chair, cup, hat, house, rabbit, tree, and watch}.

\subsubsection{Task Procedure} 
Each participant produced a total of 12 drawings, one for each object category.
Six of these drawings were cued using a category label and the other six using a photo of one of 3 typical exemplars from that category (Fig.~\ref{photodraw_task}). 
On label-cue trials, participants were instructed to ``\textit{make a drawing that would help someone else looking only at your drawing guess which word you were prompted with}'' corresponding to a category goal.
On photo-cue trials, participants were instructed to ``\textit{make a drawing that would help someone else looking only at your drawing guess which image you were prompted with, out of a lineup containing other similar images}'' corresponding to a basic-level goal.

Participants used their cursor to draw in black ink on a digital canvas (canvas: $300\times300$px; stroke width: 5px).
Each stroke was rendered in real-time on the participant’s screen as they drew and could not be deleted once drawn.
Both the label and photo cues were onscreen throughout the entire trial and participants could take as long as they wished to complete their drawing (Fig.~\ref{photodraw_data}A). 

The assignment of cue type to object category was randomized across participants, as was the order in which the object categories were displayed.
At the end of the session, participants were prompted to complete a survey in which they were asked to optionally provide the following information: sex, age, drawing device, and self-reported drawing skill. 


\subsubsection{Measuring semantic information in drawings}

This study sought to evaluate potential differences in the \textit{semantic} information conveyed by photo-cue and label-cue drawings, which can in principle be dissociable from their low-level image properties \citep{FanYaminsTurkBrowne2018}.
Measuring the semantic content in a drawing that determines its recognizability, however, requires a principled approach for encoding its high-level visual properties.
Here we leverage prior work validating the use of deep convolutional neural network (CNN) models to encode such properties in drawings \citep{FanYaminsTurkBrowne2018}.

\begin{figure*}[hbtp]
\centering
\includegraphics[width=.97\linewidth]{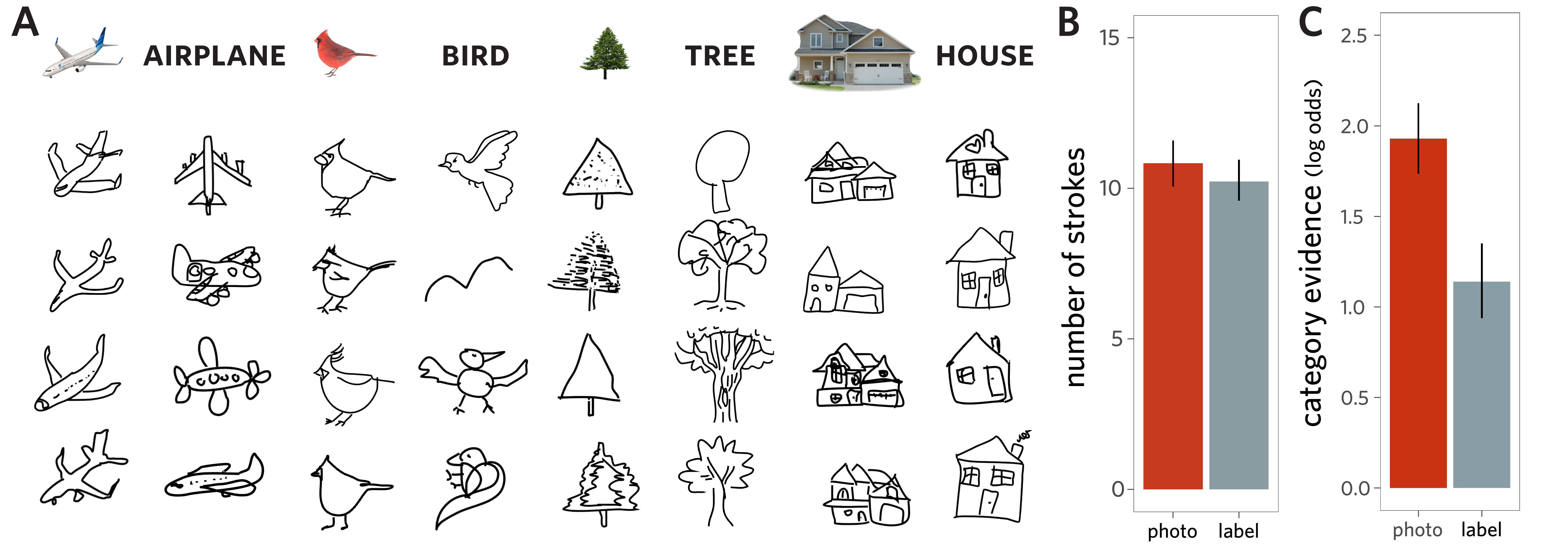} 
\caption{Study 1 Results. A: Example drawings. B: Number of strokes per drawing in the photo-cue and label-cue conditions. C: Category evidence assigned to the target category by classifier. Error bars represent 95\% bootstrap confidence intervals.}
\label{photodraw_data}
\end{figure*}

Specifically, we used VGG-19 \citep{simonyan2014very} trained to categorize objects in photos from the Imagenet database \citep{simonyan2014very} to extract high-level feature-vector representations of each sketch. 
Each 4096-dimensional feature vector reflected VGG-19 activations to drawings in the second fully-connected layer of the network (i.e., \texttt{fc6}). 
To extract the degree to which each drawing expressed the target concept, we applied a 12-way logistic classifier with L2 regularization, using 5-fold cross-validation, to predict the category label for each drawn concept.
Because this type of classifier assigns a probability value to each object, it can be used to evaluate the strength of evidence for each category contained in each drawing. 
We then used these probabilities to derive a measure that quantifies the relative evidence for the cued category compared to the others. 
Specifically, we define \textit{category evidence} to be the logodds ratio between the cued category and all other categories (Fig.~\ref{photodraw_data}C).


\subsection{Results}

\subsubsection{Drawings cued by typical exemplars are more recognizable than those cued by category labels alone}

To analyze differences in category evidence between conditions, we fit a linear mixed-effects model that included condition (i.e. photo vs. text) as a predictor, as well as random intercepts for participants and item. 
In this and subsequent statistical analyses, the best-performing linear mixed-effects model was identified using nested model comparison. 
We found that photo-cue drawings contained more category evidence than label-cue drawings (photo: 1.86, label: 0.978, $b = -0.875$, $t = -2.86$, $p = 7.08$e$-03$), suggesting that the availability of a photo of a typical exemplar may have improved participants' ability to include category-diagnostic features in their drawing. 

\subsubsection{Drawings cued by typical exemplars contain similar amounts of detail to those cued by category labels alone}

On photo-cue trials, participants had persistent access to a visual reminder of how a typical exemplar in the cued category looked. 
One potential explanation for their greater recognizability is that participants leveraged this additional information to spend more time on each trial producing drawings of greater detail. 
To test this possibility, we analyzed the number of strokes and the amount of time participants used to produce each drawing by fitting a linear mixed-effects model that included condition as a predictor (i.e., photo-cue vs. label-cue), as well as random slopes and intercepts for participants, and random intercepts for each item (i.e., the photo or the label). 
Neither analysis revealed reliable differences between conditions: participants used a similar number of strokes (photo: 10.2 strokes, label: 10.8 strokes, $b = -0.675$, $t = -0.502$, $p = 0.618$) and spent a similar amount of time (photo: 31500 ms, label: 25600 ms, $b = -5960$, $t = -1.75$, $p = 0.0876$; Fig.~\ref{photodraw_data}B) on each drawing. 
These results suggest that despite having additional visual information available on photo-cue trials, participants expended similar amounts of effort producing drawings in both conditions. 



\section{Study 2: Disentangling the contributions of sensory information, goals, and typicality}


In Study 1, we found that drawings produced while viewing a typical exemplar contained more semantically relevant information about the cued category. 
These results seem to suggest that photos generally provide useful reminders to participants of the category-diagnostic properties of objects.

However, two confounds complicate this interpretation: \textit{First}, participants cued with a photo were also instructed to produce drawings that would be discriminable at the \textit{exemplar} level, while participants cued with a label were instructed to produce drawings that would be recognizable at the \textit{category} level. 
Thus it is not clear whether the differences we observed are primarily due to the availability of sensory information (i.e., photo vs. label) or to the representational goals (i.e., to draw a category or exemplar) participants had.

\textit{Second}, the 36 photo-cues in Study 1 were all highly prototypical and perceptually similar to one another.
Thus it is not clear whether participants were more successful in producing more easily classifiable drawings on photo-cue trials due to the availability of sensory information \textit{per se}, or to low image variation, reflecting the prototypicality of these exemplars. 
To address these methodological limitations, the goal of Study 2 was to independently manipulate sensory information and representational goals, as well as  
test an expanded set of categories, each containing a larger and more heterogeneous set of exemplar images.
\subsection{Drawing Task}

\subsubsection{Participants}
We recruited 384 participants (128 female, 25.9 years) to participate in our study via Prolific.
Each participant received \$6.00 for their participation (approx. \$12/hr).
We did not exclude data from any participant, as none met our pre-registered exclusion criteria. 

\begin{figure}[htbp]
\centering
\includegraphics[width=.84\linewidth]{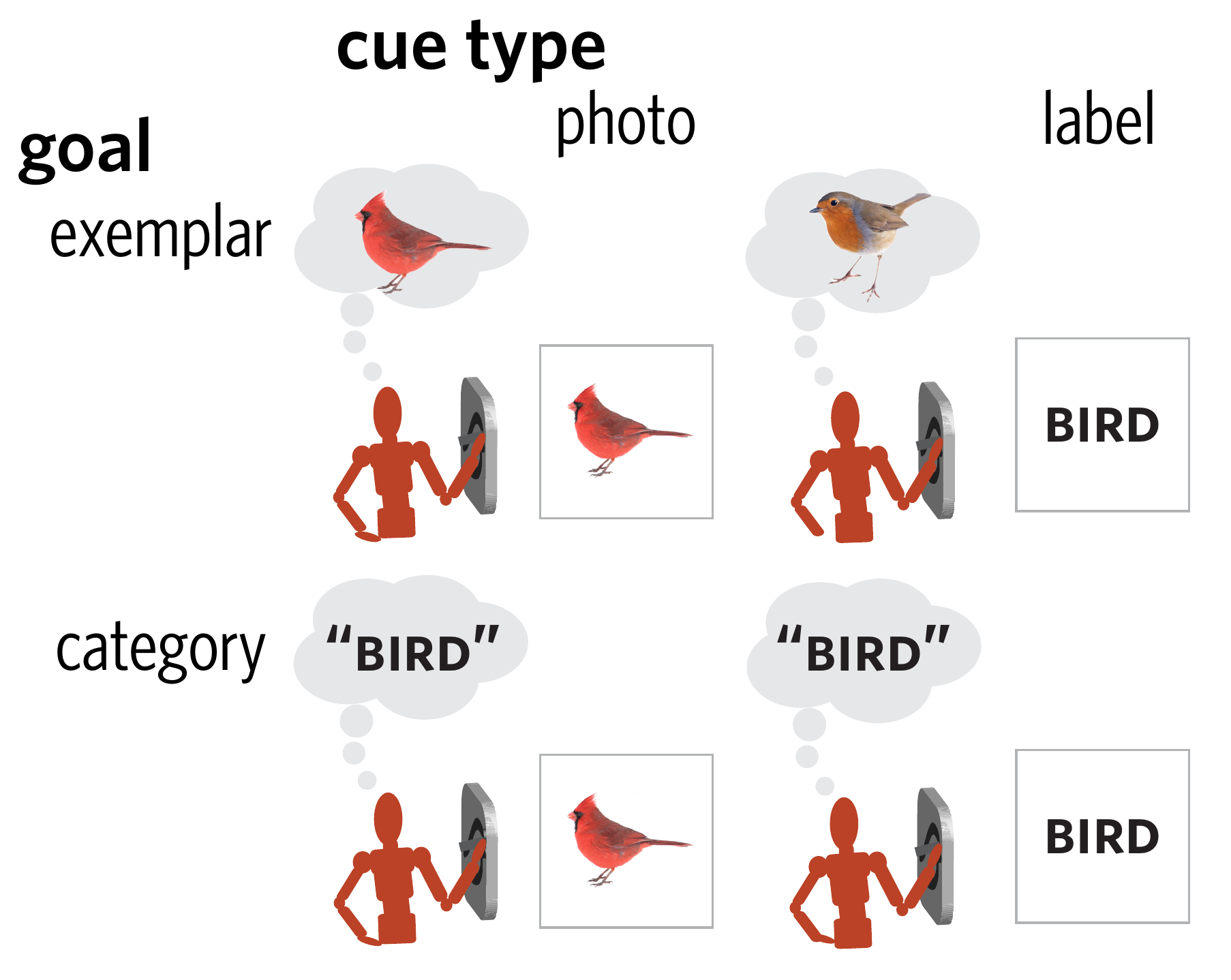}
\caption{Study 2 Task Procedure. Participants were randomly assigned to one of two goal conditions (i.e., exemplar, category) and one of two cue-type conditions (i.e., photo, label).}
\label{photodraw2x2_task}
\end{figure}

\subsubsection{Stimuli}

We included 32 basic-level categories: \textit{airplane, ape, axe, blimp, bread, butterfly, car (sedan), castle, cat, cup, elephant, fish, flower, hat, hotdog, jack-o-lantern, jellyfish, kangaroo, lion, motorcycle, mushroom, piano, raccoon, ray, saw, scorpion, skyscraper, snake, squirrel, tree, windmill, and window}.
Each category contained 32 exemplars selected from the photographs included in the Sketchy dataset \citep{sangkloy2016sketchy}.
These categories were selected to span a wide range of familiar concepts and were balanced with respect to animacy, size, familiarity, and artificiality. 
Moreover, the images \textit{within} each category were selected to vary with respect to both category-orthogonal properties (e.g., pose, viewpoint) as well as category-relevant properties (e.g., typicality). 



\subsubsection{Task Procedure}

We independently manipulated sensory information and representational goals across participants, such that each participant was pseudorandomly assigned to a cue type (i.e., photo, label) $\times$ goal (i.e., exemplar, category) condition (Fig.~\ref{photodraw2x2_task}; N=96 participants per condition).

In the photo-cue $\times$ exemplar-goal condition, participants were instructed to: ``make a drawing that would help someone else looking only at your drawing guess which image you were prompted with, out of a lineup containing other similar images.'' \footnote{These were the same instructions that photo-cue participants received in Study 1.}
In the label-cue $\times$ category-goal condition, participants were instructed to: ``make a drawing that would help someone else looking only at your drawing guess which word you were prompted with.'' \footnote{These were the same instructions that label-cue participants received in Study 1.}
In the photo-cue $\times$ category-goal condition, participants were instructed to: ``make a drawing that is recognizable, but not one that could be matched to the image I was shown.''
In the label-cue $\times$ exemplar-goal condition, participants were instructed to visualize and ``draw a \textit{specific} object, rather than a general object category.''

Each participant in Study 2 made drawings of 32 objects, with one drawing per category. 
To equate the total amount of preparation time participants in all four groups had before beginning their drawing, the cue was always presented for 8 seconds and then removed before participants could begin their drawing. 
\footnote{As a consequence, Study 2 participants who were cued with a photo \textit{did not} have persistent visual access to this image while producing their drawing, while Study 1 participants did.} 
The sequence in which categories appeared across trials was randomized across participants, but the number of times a given photo was presented was balanced, such that each photo served as the cue 3 times in Study 2. 
The resulting dataset contained 12,288 sketches.

\subsection{Measuring image typicality}

Given the greater variability between exemplars within each category, we sought to investigate potential relationships between the semantic properties of each photo --- namely, how prototypical it was --- and the properties of the resulting drawing. 
Towards this end, we crowdsourced typicality ratings for each photo. 

\subsubsection{Participants}
88 participants (42 male, 29.2 years) were recruited via Prolific. 
Each participant received \$3.00 for their participation in the ~15-minute study (approx. \$12/hr). 
Data from 8 participants who did not meet our exclusion criteria\footnote{Data from an entire session were excluded if at least 4 of 8 catch trials were failed, the same option was chosen 8 times in a row, twice in the session, or trials were rated at random (defined by abnormally low correlation with other raters).} were excluded from further analyses. 

\subsubsection{Task procedure}
Each participant was presented with the prompt (\textit{"How well does this picture fit your idea or image of the category?"}), a series of 128 images, and was asked to provide typicality judgments on a 5-point Likert scale: "Not at all", "Somewhat", "Moderately", "Very", and "Extremely". 
In each session, there were 4 images from each of the 32 categories.
This study yielded 10,240 ratings, such that each photo was rated 10 times.

\subsection{Results}

\begin{figure*}[t]
\centering
\includegraphics[width=.99\linewidth]{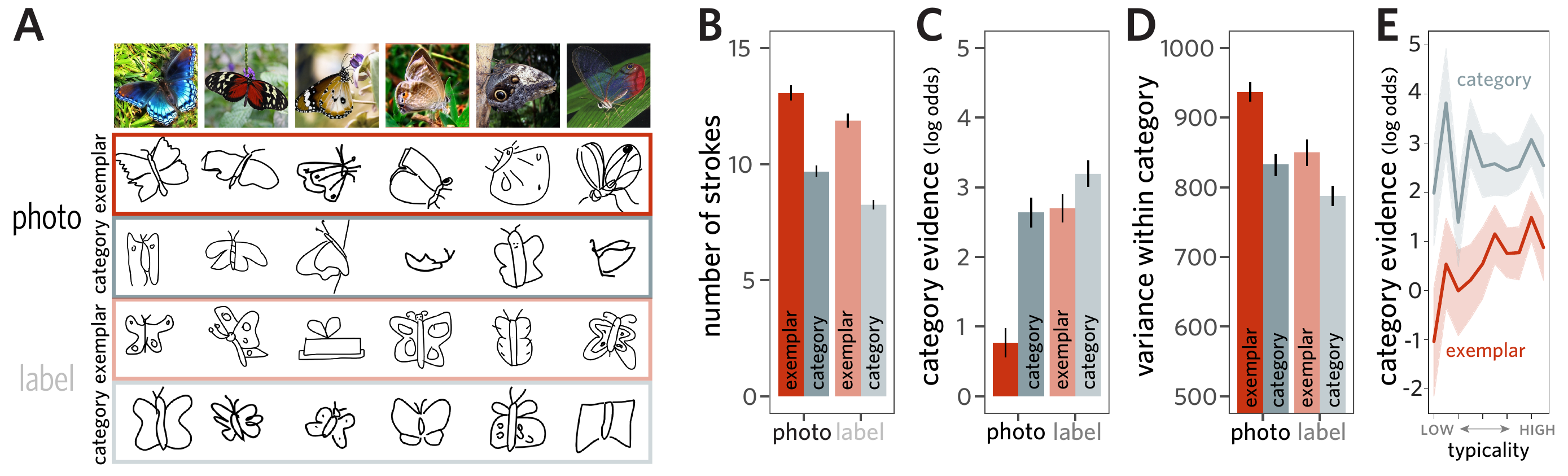}
\caption{Study 2 Results. A: Example drawings from each condition. B: Number of strokes per drawing for each condition. C: Category-level evidence across conditions. D: Variance of feature vectors for drawings within-category, for each condition. E: Relationship between category evidence and typicality among photo-cue drawings. Error bars represent 95\% bootstrap confidence intervals.}
\label{photodraw2x2_results}
\end{figure*}

\subsubsection{Differences in the amount of detail and effort between groups}

Given the results of Study 1, we did not strongly expect groups would differ in the amount of detail and effort participants would expend during drawing production in Study 2. 
Nevertheless, to evaluate any potential differences revealed by this larger dataset, we again analyzed how much time and how many strokes participants used to make each drawing. 

To test this possibility, we analyzed the number of strokes and the amount of time participants used to produce each drawing using a series of nested model comparisons among linear mixed-effects models varying in complexity.
We found that the best-performing statistical model contained fixed effects for cue type and goal (but no interaction between them), as well as random intercepts for participant and category. 
Using this model to predict the number of strokes participants used, we found a reliable main effect of representational goal (exemplar: 13.6 strokes, category: 9.83 strokes, $b = 3.48$, $t = 9.32$, $p<2$e$-16$), such that participants intending to draw specific exemplars produced more detailed drawings than those intending to draw the basic-level category, regardless of cue type (Fig.~\ref{photodraw2x2_results}B). 
The same model also revealed a main effect of cue type (photo: 12.5 strokes, label: 10.9 strokes, $b = -1.33$, $t = -3.55$, $p = 4.31$e$-04$), such that participants who were cued with a photo used more strokes than those cued with a label, regardless of their representational goal. 
When analyzing the amount of time participants spent drawing, we found a reliable main effect of representational goal (exemplar: 15.3s, category: 18.7s, $b = 3.38$, $t = 4.96$, $p = 1.09$e$-06$) but not of cue type (photo: 17.0s, label: 17.1s, $b = -0.0750$, $t = -0.112$, $p=0.911$).
Taken together, these results provide converging evidence for contributions of representational goal and cue type on the amount of effort and detail participants invest when producing their drawings.

\subsubsection{Differences in category evidence between groups}

Based on Study 1, we predicted that there would be some differences in the amount of category evidence contained by drawings from each condition, but it was not yet clear whether differences would be primarily driven by cue type, goal, or both. 
As in the previous section, we used nested model comparisons to identify the best-performing model specification, which included fixed effects for cue type, goal, their interaction, as well as self-reported drawing skill, with random intercepts for participants and category.
Using this model, we found a main effect of cue type (photo: 1.72, label: 2.90; $b = 0.554$, $t = 2.24$, $p = 0.0256$), such that label-cued drawings were actually \textit{more classifiable} than those cued by a photo, diverging from the results we obtained in Study 1 (Fig.~\ref{photodraw2x2_results}C). 
Moreover, we found a main effect of representational goal (exemplar: 1.72, category: 2.89, $b = -1.84$, $t = -7.43$, $p = 7.64$e$-13$), such that drawings intended to convey a category were more recognizable than those intended to portray a specific exemplar. 
The interaction between cue type and goal was also reliable ($b = 1.27$, $t = 3.63$, $p = 3.24$e$-04$), reflecting a larger effect of goal within the photo-cue condition. 
Finally, we observed that self-reported drawing skill had a small positive effect on drawing recognizability ($b = 0.153$, $t = 2.46$, $p = 0.0145$). 
Together, these findings provide support for the hypothesis that drawings intended to portray a specific exemplar, especially one that was recently seen, do not contain the same semantic information as drawings intended to communicate about a basic-level category.
Specifically, the additional detail that these exemplar drawings contain not only fails to enhance their recognizability but if anything, reduces their ability to evoke the cued category.  

\subsubsection{Effect of photo typicality on category evidence}
How might these results be reconciled with those obtained in Study 1? 
One of the most salient differences between the photos used in each study was how prototypical they were judged to be: while the 36 photos used in Study 1 were all maximally prototypical, the 1024 photos used in this experiment exhibited substantial and realistic variability in their visual properties (e.g., viewpoint, lighting, clutter, occlusion, size), such that none of these photos were nearly as canonical in appearance as the cues used in Study 1. 

Thus one plausible hypothesis is that the typicality of the photo cue may modulate how much category-diagnostic information participants include in their drawings, such that being cued with a less typical exemplar leads to \textit{less} recognizable drawings while being cued with a more typical exemplar leads to \textit{more} recognizable drawings, at least at the basic level. 
To evaluate this hypothesis, we fit a linear mixed-effects model predicting category evidence for photo-cue drawings only, including fixed effects for typicality, goal, their interaction, and subjective skill, as well as random intercepts for participants, category, and item (i.e., the specific photo).

In support of this hypothesis, we found that the photo-cue typicality was positively related to the amount of category evidence contained in participants' drawings ($b = 0.624$, $t = 3.45$, $p = 5.81$e$-04$; Fig.~\ref{photodraw2x2_results}E). 
Moreover, we found an interaction between the photo-cue typicality and representational goals ($b = 0.747$, $t = 3.55$, $p = 3.94$e$-04$), indicating that the positive relationship between photo-cue typicality and category evidence was stronger when participants intended to draw that particular exemplar.

\subsubsection{Estimating variability between drawings}

The fact that the additional detail participants had included in photo-cue exemplar drawings did not make them more recognizable at the category level raises the question: what are the consequences of including this extra information on the semantic properties of these drawings? 
Perhaps these drawings are characterized by a greater degree of exemplar-level discriminability, such that they are easier to tell apart from one another, even if they are not as easy to identify at the basic level as their label-cued category drawing counterparts. 

To explore this possibility, we conducted an exploratory analysis of how distinguishable drawings within a given category were, for each condition. 
Insofar as drawings that are intended to communicate about a concrete, specific exemplar are indeed more discriminable, we would predict the visual variability among drawings to be larger for photo-cued exemplar drawings than for category drawings. 
To test this prediction, we used Euclidean distance between feature vectors to compute the \textit{variance} over the set of feature vectors from each category, separately for each condition.

We found that indeed feature variability was greater overall for photo-cued drawings than for label-cued drawings (photo: 885, label: 819, $b = 65.8$, $t = -3.98$, $p = 1.17$e$-04$; Fig.~\ref{photodraw2x2_results}D). 
Moreover, feature variability was also greater for exemplar drawings than category drawings (exemplar: 893, category: 810, $b = 83.3$, $t = 9.15$, $p = 1.43$e$-15$).
Finally, we also found a reliable interaction between cue type and goal, such that the gap in feature variability for exemplar vs. category drawings was larger for the photo-cue condition than in the label-cue condition ($b = 41.3$, $t = -2.57$, $p = 0.0114$). 
Taken together, these results provide support for the notion that drawings that are intended to convey specific, concrete meanings are also more discriminable from one another than drawings intended to convey more abstract, categorical meanings.

\section{Discussion}

In this paper, we investigated the cognitive and task constraints that underlie our ability to produce drawings of object concepts at different levels of abstraction. 
This paper reports the results of both a smaller-scale exploratory study and a larger-scale follow-up study that evaluated the impact of immediate sensory inputs and representational goals on people's ability to include semantically relevant information about category membership in the drawings they produced. 
Data from Study 1 initially suggested that concurrent visual access to a photograph of an object helped people include more category-diagnostic information in their drawing than they otherwise would. 
However, data from Study 2, which was both more highly-powered and better-controlled, provided a more nuanced picture of how sensory information influences the semantic information contained in the resulting drawing: more typical photos tend to elicit more recognizable drawings than less typical ones. 
Moreover, we found that photo-cued drawings that were intended to depict that exemplar were among the least recognizable (at the category level), suggesting a dissociation between how drawings communicate more abstract vs. more specific meanings. 

Here our analyses assume that using features extracted by a convolutional neural network provides a good approximation to human sketch recognition, consistent with prior work \citep{FanYaminsTurkBrowne2018}. 
However, in ongoing work, we intend to directly validate this assumption in the current dataset by also obtaining human sketch recognition judgments. 
Another limitation of our current model-based analyses is that the architecture we used is optimized for capturing category-level information but not as well suited to representing fine-grained visual differences between exemplars. 
Thus future work seeking to more fully characterize semantic information in drawings may benefit from using architectures trained to resolve such fine-grained distinctions via techniques such as instance discrimination \citep{Wu2018,zhuang2021unsupervised}.

In this work, our study asks participants to draw real-world objects, which we assume to be strongly dependent on pre-existing knowledge.
Another promising direction for future research is to use novel objects without pre-existing associations to gain further insight into how people draw what they perceive even in the absence of verbalizable, semantic knowledge. 
More broadly, our findings highlight the value of using such open-ended production tasks to gain insight into the content and structure of conceptual knowledge.

\section{Acknowledgments}
We thank Bria Long, Mike Frank, Xuanchen Lu, and members of the Cognitive Tools Lab at UC San Diego for helpful discussion.
This work was supported by a Halıcıoğlu Data Science Undergraduate Scholarship and the UCSD Chancellor's Research Scholarship awarded to J.Y., as well as by NSF CAREER Award \#2047191 to J.E.F.

\vspace{2em}
\fbox{\parbox[b][][c]{7.3cm}{\centering {All code and materials available at: \\
\href{https://github.com/cogtoolslab/photodraw}{\url{https://github.com/cogtoolslab/photodraw_cogsci2021}}
}}}
\vspace{2em} \noindent

\setlength{\bibleftmargin}{.125in}
\setlength{\bibindent}{-\bibleftmargin}

\bibliography{references}

\begin{thebibliography}{}

\bibitem[\protect\astroncite{Bainbridge et~al.}{2019}]{bainbridge2019drawings}
Bainbridge, W.~A., Hall, E.~H., and Baker, C.~I. (2019).
\newblock Drawings of real-world scenes during free recall reveal detailed
  object and spatial information in memory.
\newblock {\em Nature Communications}, 10(1):1--13.

\bibitem[\protect\astroncite{Bozeat et~al.}{2003}]{Bozeat:2003hk}
Bozeat, S., Lambon~Ralph, M.~A., Graham, K.~S., Patterson, K., Wilkin, H.,
  Rowland, J., Rogers, T.~T., and Hodges, J.~R. (2003).
\newblock {A duck with four legs: Investigating the structure of conceptual
  knowledge using picture drawing in semantic dementia}.
\newblock {\em Cognitive Neuropsychology}, 20(1):27--47.

\bibitem[\protect\astroncite{Chen and Goedert}{2012}]{chen2012clock}
Chen, P. and Goedert, K.~M. (2012).
\newblock Clock drawing in spatial neglect: A comprehensive analysis of clock
  perimeter, placement, and accuracy.
\newblock {\em Journal of Neuropsychology}, 6(2):270--289.

\bibitem[\protect\astroncite{Clottes}{2008}]{clottes2008cave}
Clottes, J. (2008).
\newblock {\em Cave Art}.
\newblock Phaidon London.

\bibitem[\protect\astroncite{Dillon}{2020}]{dillon2020rooms}
Dillon, M.~R. (2020).
\newblock Rooms without walls: Young children draw objects but not layouts.
\newblock {\em Journal of Experimental Psychology: General}.

\bibitem[\protect\astroncite{Fan et~al.}{2018}]{FanYaminsTurkBrowne2018}
Fan, J., Yamins, D., and Turk-Browne, N. (2018).
\newblock Common object representations for visual production and recognition.
\newblock {\em Cognitive Science}, 42:2670--2698.

\bibitem[\protect\astroncite{Fan et~al.}{2020}]{fan2020pragmatic}
Fan, J.~E., Hawkins, R.~D., Wu, M., and Goodman, N.~D. (2020).
\newblock Pragmatic inference and visual abstraction enable contextual
  flexibility during visual communication.
\newblock {\em Computational Brain \& Behavior}, 3(1):86--101.

\bibitem[\protect\astroncite{Fiorella and Zhang}{2018}]{fiorella2018drawing}
Fiorella, L. and Zhang, Q. (2018).
\newblock Drawing boundary conditions for learning by drawing.
\newblock {\em Educational Psychology Review}, 30(3):1115--1137.

\bibitem[\protect\astroncite{Gombrich}{1989}]{gombrich1989story}
Gombrich, E. (1989).
\newblock {\em The story of art}.
\newblock Phaidon Press, Ltd.

\bibitem[\protect\astroncite{Hawkins et~al.}{2019}]{hawkins2019disentangling}
Hawkins, R.~X., Sano, M., Goodman, N.~D., and Fan, J.~W. (2019).
\newblock Disentangling contributions of visual information and interaction
  history in the formation of graphical conventions.
\newblock In {\em CogSci}, pages 415--421.

\bibitem[\protect\astroncite{Karmiloff-Smith}{1990}]{KarmiloffSmith:1990ty}
Karmiloff-Smith, A. (1990).
\newblock {Constraints on representational change: Evidence from children's
  drawing}.
\newblock {\em Cognition}, 34(1):57--83.

\bibitem[\protect\astroncite{Long et~al.}{2019}]{Long2019}
Long, B., Fan, J.~E., Chai, Z., and Frank, M.~C. (2019).
\newblock Developmental changes in the ability to draw distinctive features of
  object categories.
\newblock {\em Journal of Vision}, 19.

\bibitem[\protect\astroncite{Lupyan and
  Thompson-Schill}{2012}]{lupyan2012evocative}
Lupyan, G. and Thompson-Schill, S.~L. (2012).
\newblock The evocative power of words: activation of concepts by verbal and
  nonverbal means.
\newblock {\em Journal of Experimental Psychology: General}, 141(1):170.

\bibitem[\protect\astroncite{Miller}{1995}]{miller1995wordnet}
Miller, G.~A. (1995).
\newblock Wordnet: a lexical database for english.
\newblock {\em Communications of the ACM}, 38(11):39--41.

\bibitem[\protect\astroncite{Minsky and Papert}{1972}]{minsky1972artificial}
Minsky, M. and Papert, S.~A. (1972).
\newblock Artificial intelligence progress report.

\bibitem[\protect\astroncite{Roberts and Wammes}{2020}]{roberts2020drawing}
Roberts, B.~R. and Wammes, J.~D. (2020).
\newblock Drawing and memory: Using visual production to alleviate concreteness
  effects.
\newblock {\em Psychonomic Bulletin \& Review}, pages 1--9.

\bibitem[\protect\astroncite{Rosch et~al.}{1976}]{rosch1976basic}
Rosch, E., Mervis, C.~B., Gray, W.~D., Johnson, D.~M., and Boyes-Braem, P.
  (1976).
\newblock Basic objects in natural categories.
\newblock {\em Cognitive Psychology}, 8(3):382--439.

\bibitem[\protect\astroncite{Sangkloy et~al.}{2016}]{sangkloy2016sketchy}
Sangkloy, P., Burnell, N., Ham, C., and Hays, J. (2016).
\newblock The sketchy database: Learning to retrieve badly drawn bunnies.
\newblock {\em ACM Transactions on Graphics (proceedings of SIGGRAPH)}.

\bibitem[\protect\astroncite{Simonyan and Zisserman}{2014}]{simonyan2014very}
Simonyan, K. and Zisserman, A. (2014).
\newblock Very deep convolutional networks for large-scale image recognition.
\newblock {\em arXiv preprint arXiv:1409.1556}.

\bibitem[\protect\astroncite{Wu et~al.}{2018}]{Wu2018}
Wu, Z., Xiong, Y., Yu, S.~X., and Lin, D. (2018).
\newblock Unsupervised feature learning via non-parametric instance
  discrimination.
\newblock In {\em Proceedings of the IEEE Conference on Computer Vision and
  Pattern Recognition}, pages 3733--3742.

\bibitem[\protect\astroncite{Xu and Tenenbaum}{2007}]{xu2007word}
Xu, F. and Tenenbaum, J.~B. (2007).
\newblock Word learning as bayesian inference.
\newblock {\em Psychological Review}, 114(2):245.

\bibitem[\protect\astroncite{Zhuang et~al.}{2021}]{zhuang2021unsupervised}
Zhuang, C., Yan, S., Nayebi, A., Schrimpf, M., Frank, M.~C., DiCarlo, J.~J.,
  and Yamins, D.~L. (2021).
\newblock Unsupervised neural network models of the ventral visual stream.
\newblock {\em Proceedings of the National Academy of Sciences}, 118(3).

\end{thebibliography}

\end{document}